\documentclass[letterpaper, 10 pt, conference]{ieeeconf}  

\IEEEoverridecommandlockouts                              

\usepackage{graphicx}
\usepackage{booktabs,caption}
\usepackage{epsfig} 
\usepackage{amsmath} 
\usepackage{amssymb}  
\usepackage{bm}
\usepackage{multicol, blindtext}
\usepackage{float}
\usepackage{subcaption}
\usepackage{siunitx}
\usepackage{xcolor}
\usepackage{hyperref}
\newcommand{\diag}{\mathop{\mathrm{diag}}}

\title{\LARGE \bf
Asynchronous Real-Time Optimization of Footstep Placement\\ and Timing in Bipedal Walking Robots
}

\author{Digby Chappell$^{*}$, Ke Wang$^{*}$, and Petar Kormushev
\thanks{$^{*}$Equal Contribution}
\thanks{All authors are with the Robot Intelligence Lab, Dyson School of Design Engineering, Faculty of Engineering, Imperial College London, SW7 2DB, United Kingdom. Contact:
        {\tt\small \{d.chappell19, k.wang17, p.kormushev\}@imperial.ac.uk%
        }
    }
\thanks{Supplementary video can be viewed at: \url{https://www.youtube.com/watch?v=Jw5_cMZvoEI}}
}
\begin{document}

\maketitle
\thispagestyle{empty}
\pagestyle{empty}

\begin{abstract}

Online footstep planning is essential for bipedal walking robots to be able to walk in the presence of disturbances. Until recently this has been achieved by only optimizing the placement of the footstep, keeping the duration of the step constant. In this paper we introduce a footstep planner capable of optimizing footstep placement and timing in real-time by asynchronously combining two optimizers, which we refer to as asynchronous real-time optimization (ARTO). The first optimizer which runs at approximately 25\,Hz, utilizes a fourth-order Runge-Kutta (RK4) method to accurately  approximate the dynamics of the linear inverted pendulum (LIP) model for bipedal walking, then uses non-linear optimization to find optimal footsteps and duration at a lower frequency. The second optimizer that runs at approximately 250\,Hz, uses analytical gradients derived from the full dynamics of the LIP model and constraint penalty terms to perform gradient descent, which finds approximately optimal footstep placement and timing at a higher frequency. By combining the two optimizers asynchronously, ARTO has the benefits of fast reactions to disturbances from the gradient descent optimizer, accurate solutions that avoid local optima from the RK4 optimizer, and increases the probability that a feasible solution will be found from the two optimizers. Experimentally, we show that ARTO is able to recover from considerably larger pushes and produces feasible solutions to larger reference velocity changes than a standard footstep location optimizer, and outperforms using just the RK4 optimizer alone.

\end{abstract}

\section{INTRODUCTION}

In order for robots to walk robustly in the presence of noise and disturbances, footstep replanning is required. In many works, this has been achieved by having fixed step timing, and replanning the location of footsteps according to the optimal control of the LIP model \cite{Motoi2009, Kajita2010}. The LIP model is a widely used approximate model of the dynamics of a bipedal walking robot; the legs are assumed to be massless, and the body is a concentrated point mass located at the robot's centre of mass (CoM) and the height is assumed to be constant \cite{Kajita2001}. Because the LIP model is governed by a second order differential equation, optimizing footsteps with respect to time produces a non-linear optimization problem which is hard to solve in real-time. Thus, real-time footstep timing adaptation work has largely been focused on approximate solutions or simplified dynamics control problems.

One popular method to simplify the control problem is to consider only the divergent component of motion (DCM) of the body. By only considering the DCM, dynamics are reduced to a first order problem, where the optimal step time to control the DCM about a nominal point can be computed with quadratic programming (QP) \cite{Khadiv2016, Khadiv2020}. DCM can also be combined with the a virtual repellent point \cite{Englsberger2013} to encode all forces on a robot's body in one single point. \cite{Englsberger2017} controls the DCM in this way by interpolating the virtual repellent point between desired locations. However, the DCM only presents a simplified version of the model dynamics, potentially limiting the control capabilities of footstep planning.

Instead, another approach is to optimize the full system dynamics, only considering the CoM state at each footstep. \cite{Kryczka2015} achieves this, interpolating between regenerated gait patterns when a distburance is detected. \cite{Ding2019} also only considers the CoM state at each step, but reformulates the nonlinear optimization problem to a sequential quadratic programming problem by writing the nonlinear terms directly as decision variables. However, although \cite{Ding2019} is fast, the decision variables, including footstep positions require references to track - this may not represent optimal behaviour.

Because the full solution of the LIP dynamics is nonlinear, the computational time of directly solving the nonlinear optimization is high \cite{hu2018comparison}. Rather than to consider the full dynamics at each footstep, a reasonable simplification is to approximately discretize the dynamics of the system. This greatly reduces the complexity of the optimization problem, but introduces a new problem of accurate discretization. In this work we consider explicit Runge-Kutta methods to approximate the solution at each footstep. Runge-Kutta methods have been used in literature to discretize nonlinear dynamics for model predictive control (MPC) problems \cite{Gros2020}, and have been found particularly useful in real time iteration (RTI) \cite{Houska2012, Yang2015}. Regarding applications to robotics, using explicit Runge-Kutta integrator methods to discretize nonlinear dynamics has been empirically shown to be fast \cite{Arbo2017}, and generalizes well to higher degree of freedom robots, as the full solution to the system dynamics is not required. In this work we discretize the motion of the LIP model into a fixed number of segments per step approximately using RK4, and utilize this to optimize footstep position and timing at a rate of approximately $25\,$\si{Hz}.

Real-time nonlinear MPC approaches focus on generating approximate solutions that are iteratively improved with each control step. Real time iteration (RTI), introduced in \cite{Diehl2002} is an online method for approximately solving MPC problems, where the solution is found by first linearising the system at its current state, then performing one quadratic programming (QP) step \cite{Diehl2005}. RTI has been shown to significantly reduce computation time when solving nonlinear MPC problems \cite{Gros2020}. Similar real-time works include the Continuation/Generalized Minimal Residual Method, which is based on interior point methods \cite{Ohtsuka2004}, and the Advanced-Step Nonlinear MPC Controller, which predicts the system state and optimal input at the next control time then adapts this solution to the actual system state \cite{Zavala2009}. Gradient-based methods also present benefits for online MPC problems; \cite{Richter2009, Richter2012} outline how the fast gradient method \cite{Nesterov1983} can handle input constraints for a linear MPC problem without significant increase in complexity. Gradient descent has previously been used to solve MPC problems \cite{Kim2003}, but has proved difficult when using the projected gradient method to handle constraints \cite{Torrisi2018}. Instead, constraints can be introduced to gradient descent as `soft' constraints - penalty terms included in the cost function to shape it such that optima satisfy constraints.

\begin{figure}[hbt]
    \centering
    \vspace{-0.3cm}
    \includegraphics[width=0.375\textwidth]{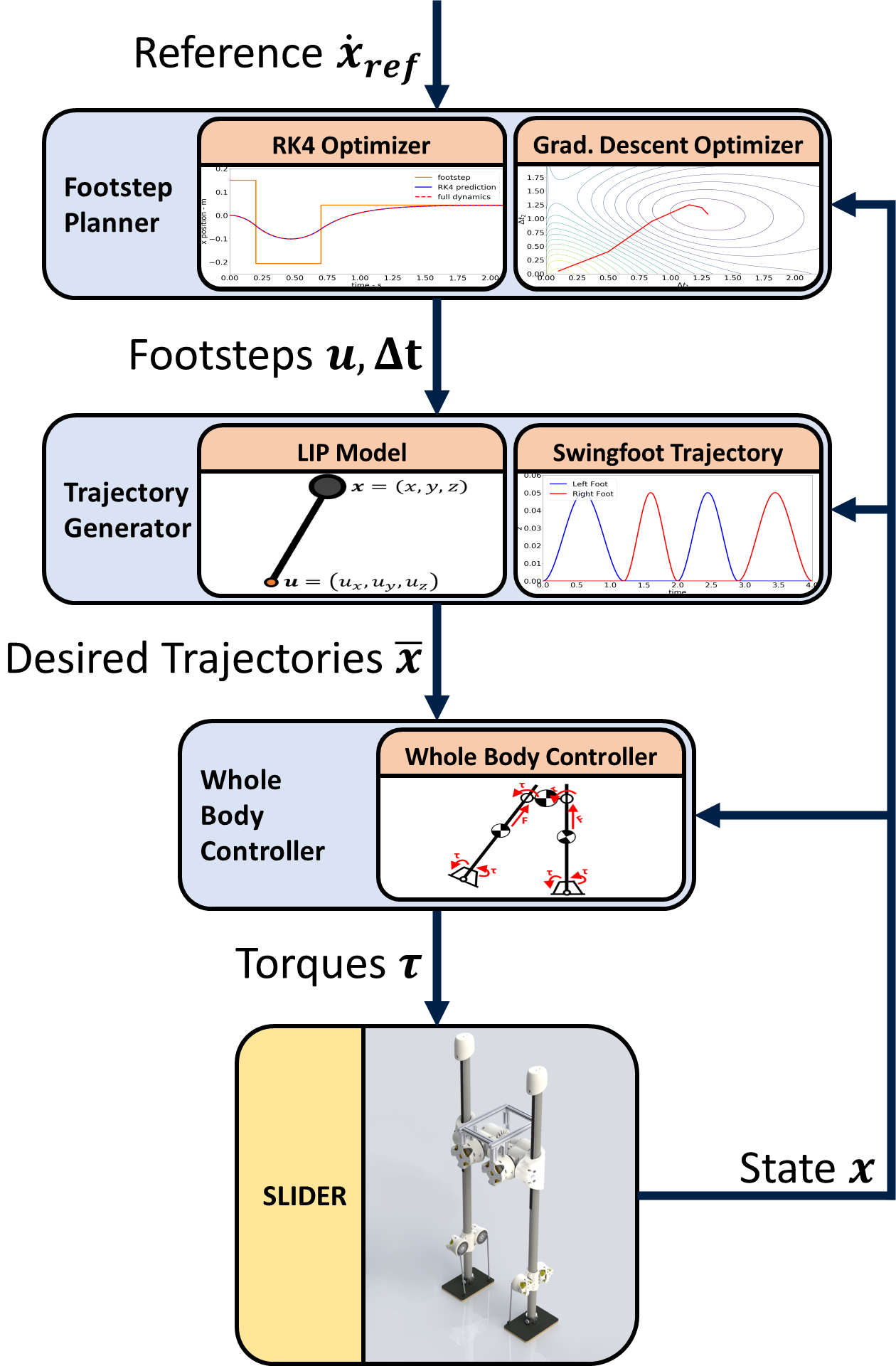}
    \caption{Proposed control diagram, including the proposed footstep planner.}    \vspace{-0.5cm}
    \label{fig:Control}
\end{figure}

In this work we investigate the use of gradient descent as a `real-time' approach to the nonlinear bipedal walking problem; a fixed number of gradient descent steps are iterated at each control step, taking the approximate solution from the previous control step as the initial solution for the next. Specific to the LIP model, we calculate these gradients analytically, allowing each gradient descent step to be computed in sub-millisecond time. However, as with many gradient-based methods, gradient descent is prone to converging on local optima. To prevent this becoming a problem, we run the RK4 optimizer in parallel at a lower frequency than gradient descent, that can then update the footstep plan with a more optimal solution. The asynchronous combination of these two optimizers is hence referred to as Asynchronous Real-Time Optimization (ARTO).

\begin{figure}[hbt]
    \centering
    \includegraphics[width=0.35\textwidth]{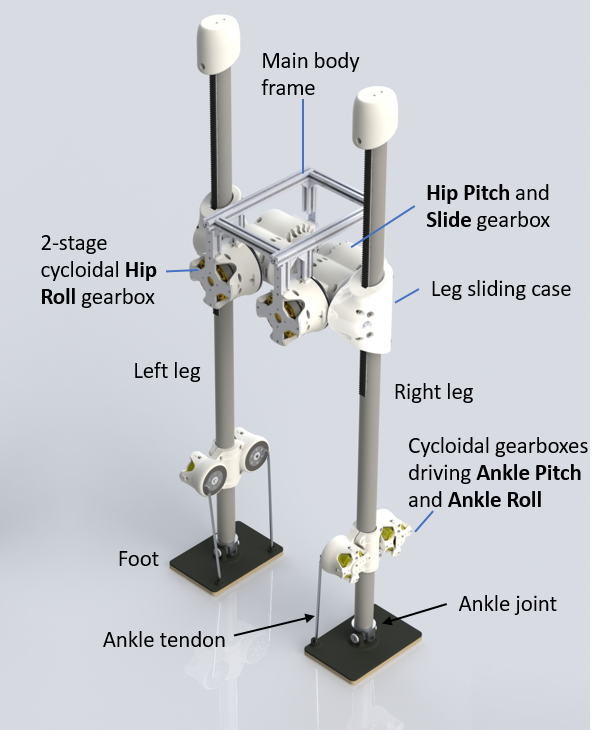}
    \caption{SLIDER walking robot platform.}
    \vspace{-0.5cm}
    \label{fig:SLIDER}
\end{figure}

ARTO combines the best of two optimizers; the non-linear optimization with RK4 dynamics produces a near-optimal solution, and gradient descent can quickly update the solution, which enables real-time planning. To validate ARTO, we test it in simulation as part of the control framework (see Fig. \ref{fig:Control}) of the walking robot SLIDER (see Fig. \ref{fig:SLIDER}). Three sets of experiments are conducted, first observing push recovery behaviour, then quantifying push recovery capabilities and reference velocity change handling capacity of the footstep planner against alternative footstep planning algorithms.


\section{Asynchronous Real-Time Optimization}
\subsection{Preliminaries}

The LIP model for bipedal walking in the sagittal plane is defined by (\ref{eq:LIP}), where the CoM has position $x$ and fixed height $h$, and $u_x$ is the position of the foot. Note that the motion in the coronal plane is defined by identical equations for $y$, as the constant height decouples the motion from the sagittal plane.
\begin{equation}
    \Ddot{x} = \frac{g}{h}(x - u_x)\label{eq:LIP}
\end{equation}
This can be equivalently written in state space form for both the sagittal and coronal motion, with $\bm{x} = [x,\, y,\, \Dot{x},\, \Dot{y}]^T$, and $\bm{u} = [u_x,\, u_y]^T$:
\begin{equation}
    \Dot{\bm{x}}
    =
    \begin{bmatrix}
        0   & 0     & 1     & 0 \\
        0   & 0     & 0     & 1 \\
        g/h & 0     & 0     & 0 \\
        0   & g/h   & 0     & 0
    \end{bmatrix}
    \bm{x}
    +
    \begin{bmatrix}
        0 & 0\\
        0 & 0\\
        -g/h & 0\\
        0 & -g/h
    \end{bmatrix}
    \bm{u}
    \label{eq:ssLIP}
\end{equation}
The full solution to (\ref{eq:LIP}) can be expressed as a hyperbolic equation (\ref{eq:sinhcosh}), or an exponential equation (\ref{eq:exp}), where $\omega=\sqrt{g/h}$. Here we define $\Delta t_k$ as the step duration, $x_k$ and $\Dot{x}_k$ as the CoM position and velocity at the beginning of footstep $k$, and $u_{x,k}$ as the position of the foot (which we assume to be fixed for the duration of the step).
\begin{equation}
    x_{k+1} = \frac{\Dot{x}_{k}}{\omega}\sinh{\omega \Delta t_k} + (x_k - u_{x,k})\cosh{\omega \Delta t_k} + u_{x,k}\label{eq:sinhcosh}
\end{equation}
\begin{equation}
    x_{k+1} = a_{k}\exp(\omega \Delta t_k) + b_{k}\exp(-\omega \Delta t_k) + u_{x,k}\label{eq:exp}
\end{equation}
Where $a_k = 0.5(x_k - u_{x,k} + \Dot{x}_k/\omega)$, and $b_k = 0.5(x_k - u_{x,k} - \Dot{x}_k/\omega)$. Note that the positive exponential term is linked to the divergent component of motion, as it diverges when $\Delta t_k$ increases. From here we define our optimization problem to control the velocity of the CoM at each footstep in the $x$ and $y$ directions in state space form, with control variables $\bm{u}$ and $\bm{\Delta t}$:
\begin{equation}
    \min_{\bm{u},\, \bm{\Delta t}} \sum_{k=1}^{N}\big((\bm{x}_k-\bm{x}_{ref})^T\bm{Q}(\bm{x}_k-\bm{x}_{ref})\big)\label{eq:cost}
\end{equation}
Where here we use $\bm{Q}=\diag(0,\, 0,\, w_x,\, w_y)$ to track velocity. $w_x$ and $w_y$ are the weightings placed on the $x$ and $y$ velocity costs, respectively. We then define the following dynamic constraint:
\begin{equation}
    \Dot{\bm{x}}_k = \bm{f}\big(\bm{x}(\bm{u}_k, \,\Delta t_k),\, \bm{u}_k\big) \label{eq:dynamic_constraints} 
\end{equation}
Then physical constraints, with $\bm{x}_{pos,k}=[x_k, y_k]^T$:
\begin{equation}
    ||\bm{x}_{pos,k} - \bm{u}_{k}||_2 \leq l_{\text{max}}
    \label{eq:current_step_constraint}
\end{equation}
\begin{equation}
    ||\bm{x}_{pos,k} - \bm{u}_{k-1}||_2 \leq l_{\text{max}}
    \label{eq:next_step_constraint}
\end{equation}
\begin{equation}
    |\bm{u}_{k, \, y} - \bm{u}_{k-1, \, y}| \geq r_{\text{foot}}
    \label{eq:foot_constraints}
\end{equation}
\begin{equation}
    c_{\text{t, k, lower}} \leq \Delta t_k \leq c_{\text{t, k, upper}}
    \label{eq:time_constraints}
\end{equation}
Where (\ref{eq:current_step_constraint}) and (\ref{eq:next_step_constraint}) prevent the leg from reaching its extension limit, $l_{\text{max}}$, when travelling on the current support foot and when placing the next support foot, respectively. (\ref{eq:foot_constraints}) prevents the next support foot from crossing over the current support foot, with a limit of $r_{\text{foot}}$. (\ref{eq:time_constraints}) places an upper and lower limit on the step duration, with the upper limit, $c_{\text{t, k, upper}}$, encouraging regular steps, and the lower limit, $c_{\text{t, k, lower}}$, also to ensure that the swing foot does not exceed its velocity limit.

As \cite{Khadiv2020} found and compared to human studies \cite{Patla2003}, it is sufficient to consider just the next step when walking normally, and two steps ahead when walking on difficult terrains or with considerable constraints. In this work we consider a prediction horizon of two footsteps ahead; optimizing two footstep positions and three footstep durations (including the current footstep duration).

\subsection{Runge-Kutta Optimization}

\begin{figure}[hbt]
    \centering
    \includegraphics[width=0.45\textwidth]{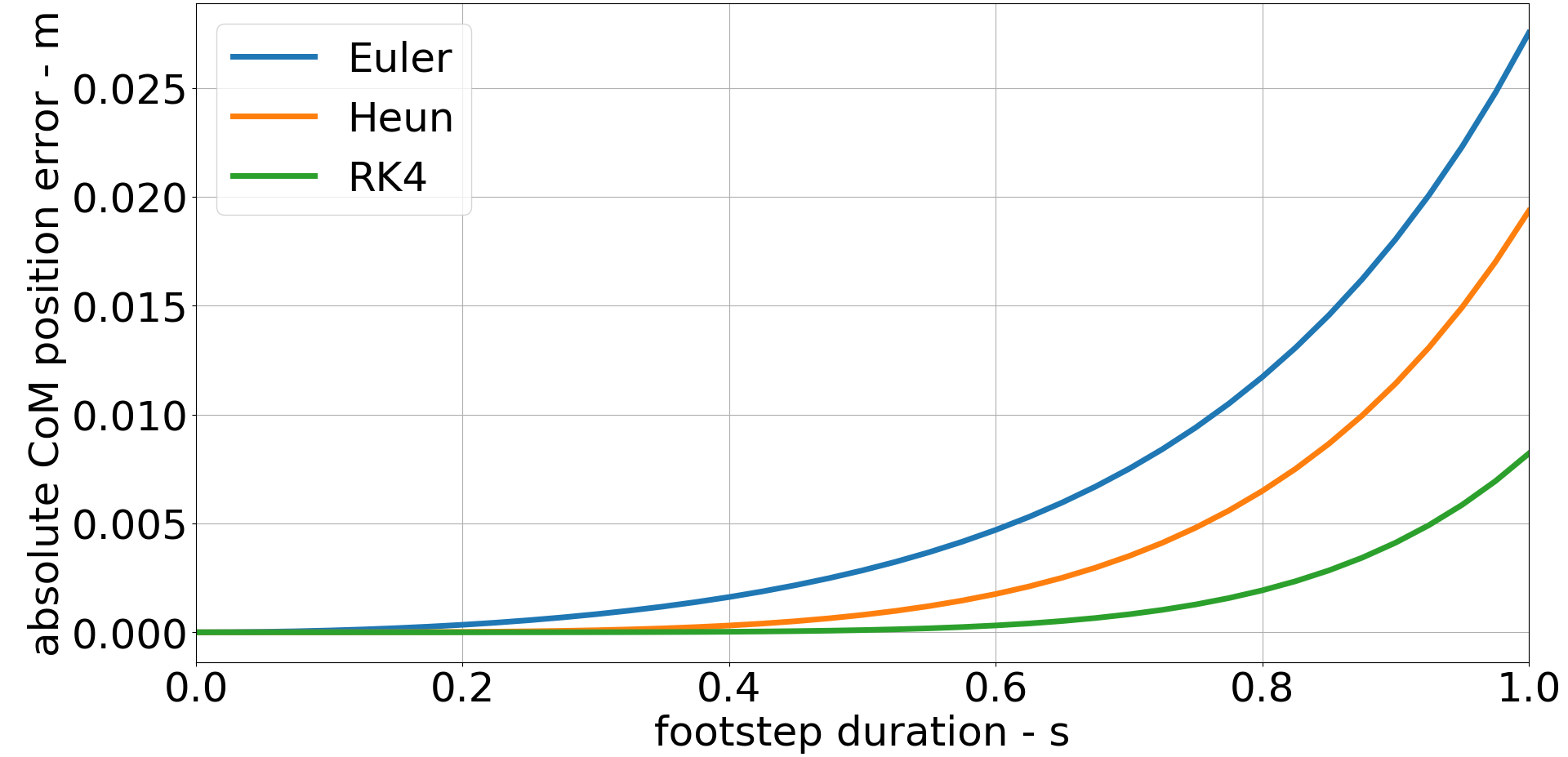}
    \includegraphics[width=0.43\textwidth]{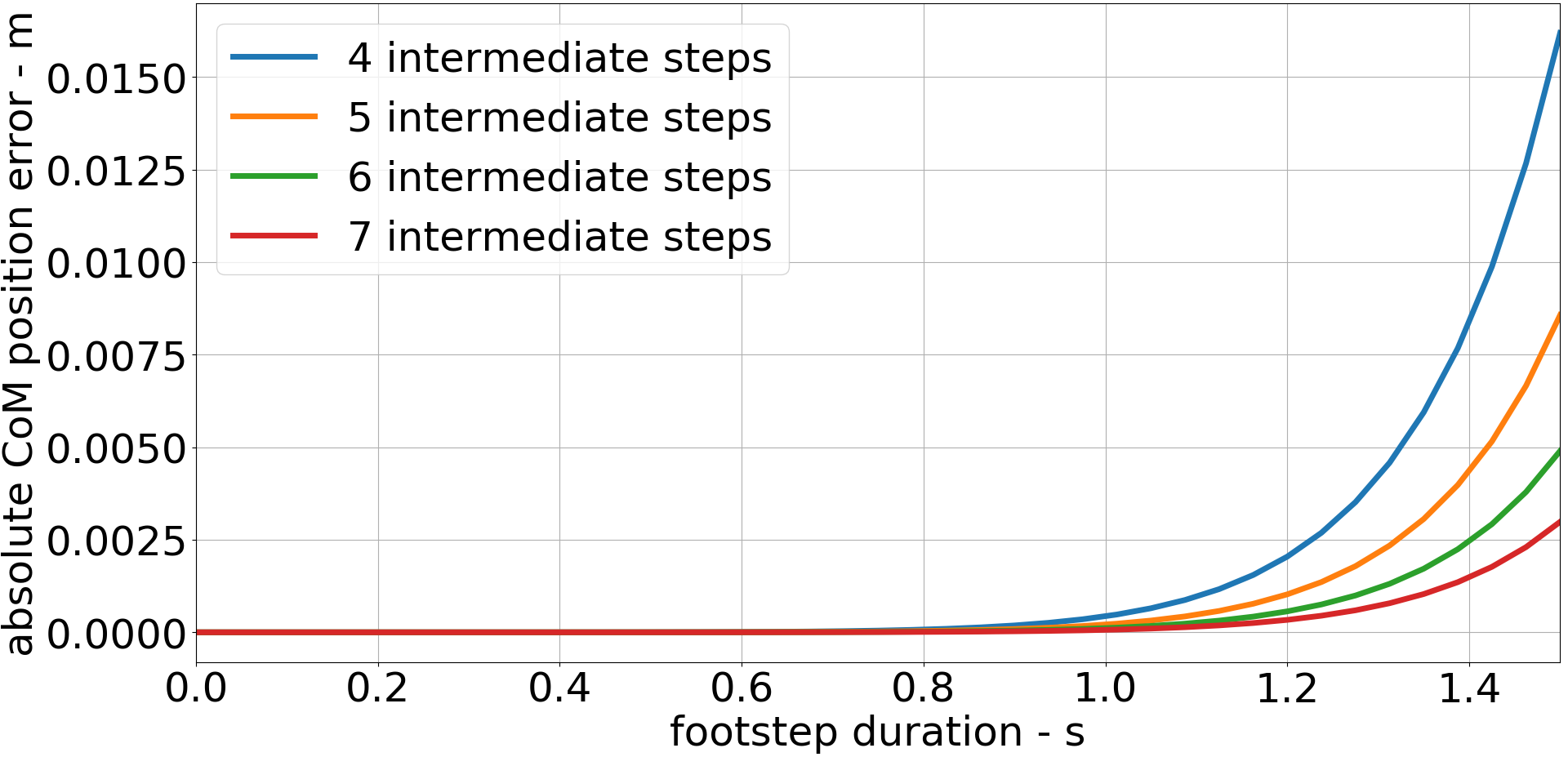}
    \caption{Top: Absolute CoM position error produced by Euler (blue), Heun (orange), and RK4 (green) numerical methods. Bottom: Absolute CoM position error produced by using 4 (red), 5 (orange), 6 (green), and 7 (red) intermediate steps.}
    \vspace{-0.3cm}
    \label{fig:errors}
\end{figure}

Explicit Runge-Kutta methods can be used to numerically integrate the LIP model dynamics over a specific step size, $\delta t$, given differential equation, $y'(t) = f(t, y(t))$. Three Runge-Kutta methods were considered: forward Euler method, Heun's method, and RK4. RK4, a fourth order Runge-Kutta method is defined as:
\begin{equation*}
    k_1 = \delta t \, f(t_k, y_{t_k})
\end{equation*}

\begin{equation*}
    k_2 = \delta t \, f(t_k + \frac{\delta t}{2}, y_{t_k} + \frac{k_1}{2})
\end{equation*}
\begin{equation*}
    k_3 = \delta t \, f(t_k + \frac{\delta t}{2}, y_{t_k} + \frac{k_2}{2})
\end{equation*}
\begin{equation*}
    k_4 = \delta t \, f(t_k + \delta t, y_{t_k} + k_3)
\end{equation*}
\begin{equation}
    y_{t_k+\delta t} = y_{t_k} + \frac{1}{6}\big( k_1 + 2k_2 + 2k_3 + k_4\big) \label{eq:rk4} 
\end{equation}
The most suitable method for approximating the dynamics of the LIP model is RK4; this is illustrated in Fig. \ref{fig:errors} for a sagittal plane LIP model initially at rest, with body height $0.6\,$m and support foot offset $1\,$mm. As shown, RK4 produces smaller positional errors than Heun's method or forward Euler, so RK4 is suitable when discretizing to a low number of intermediate points.

To simplify the nonlinear optimization problem posed by the LIP dynamics but maintain accuracy of approximation, RK4 is used to discretize each footstep into $n$ intermediate steps. Defining $f^{\circ n}(x)$ as the composition of function $f$ with itself $n$ times, that is:
\begin{equation}
    f^{\circ 2}(x) = f\big( f(x) \big)
\end{equation}
Then letting $\delta t = (t_{k+1} - t_k)/n$, and using (\ref{eq:rk4}), the CoM state one footstep after step $k$ can be written:
\begin{equation}
    \bm{x}_{k+1} = f^{\circ n}_{rk4}(\bm{x}_{k}, \, \bm{u}_{k}, \, \delta t)
    \label{eq:rk4_one_step}
\end{equation}
The number of intermediate steps to discretize each footstep was chosen as a trade-off between computational complexity and prediction error minimization. As shown in Fig. \ref{fig:errors} for a sagittal plane LIP model initially at rest, with body height $0.6\,$m and support foot offset $1\,$mm, increasing the number of intermediate steps produces diminishing returns. Six intermediate steps was selected to reduce prediction error while maintaining low complexity.

The optimization problem using RK4 to discretize the dynamics can therefore be written as previous, replacing (\ref{eq:dynamic_constraints}) with:

\begin{equation*}
    \bm{x}_{k+1} = f^{\circ 6}_{rk4}(\bm{x}_{k}, \, \bm{u}_{k}, \, \Delta t_k / 6)
\end{equation*}

We refer to the optimizer that performs this non-linear optimization with approximate dynamics as the RK4 optimizer.

\subsection{Gradient Descent Optimization}

Gradient descent methods, using analytically calculated gradients, can be used to find approximate minima in the footstep planning cost function very quickly. However, gradient methods often have difficulty handling constraints. To remedy this, we add penalty functions to our cost function as a form of `soft' constraint. To convert an inequality constraint to a soft penalty, the following formulation is used, where $w_i$ is a weighting to be tuned:
\begin{equation*}
    g_i(\bm{x},\, \bm{u}, \, \Delta t) \leq c_i
\end{equation*}
\begin{equation}
    p_i(\bm{x},\, \bm{u}, \, \Delta t) = w_i\exp\big( g_i(\bm{x},\, \bm{u}, \, \Delta t) / c_i \big)
    \label{eq:soft_constraint}
\end{equation}
Giving a full cost function, $J$, with $M$ constraints:
\begin{equation}
    J_p = J + \sum_{i=1}^{M}p_i
\end{equation}

Equality constraints can be directly inserted to the cost function formulation, where analytical gradients can then be calculated with respect to footstep position in the $x$ direction:
\begin{equation}
    \frac{\partial J_p}{\partial u_{x,i}} = 2w_x\sum_{k=1}^{N}\frac{\partial \dot{x}_k}{\partial u_{x,i}}(\dot{x}_k - \dot{x}_{ref}) + \frac{\partial}{\partial u_x}\sum_{i=1}^{M}p_i
    \label{eq:grad_position}
\end{equation}
With similar formulation in the $y$ direction. The gradient with respect to footstep duration can be written:
\begin{equation}
    \frac{\partial J_p}{\partial \Delta t_{i}} = 2w_x\sum_{k=1}^{N}\frac{\partial \bm{x}_k}{\partial \Delta t_{i}}^T\bm{Q}(\bm{x}_k-\bm{x}_{ref}) + \frac{\partial}{\partial \Delta t_{i}}\sum_{i=1}^{M}p_i
    \label{eq:time_position}
\end{equation}
Because the cost function is quadratic and constraints formulated as exponential penalties, these gradients are simple to compute, and gradient descent steps can be executed quickly. Expressions for $\frac{\partial x_k}{\partial \Delta u_{x,i}}$, $\frac{\partial \dot{x}_k}{\partial \Delta u_{x,i}}$, $\frac{\partial x_k}{\partial \Delta t_{i}}$, and $\frac{\partial \dot{x}_k}{\partial \Delta t_{i}}$ can be found in Appendix A.

\subsection{Asynchronous Optimization}

\begin{figure*}[hbt]
    \centering
    \vspace{0.5cm}
    \includegraphics[width=0.7\textwidth]{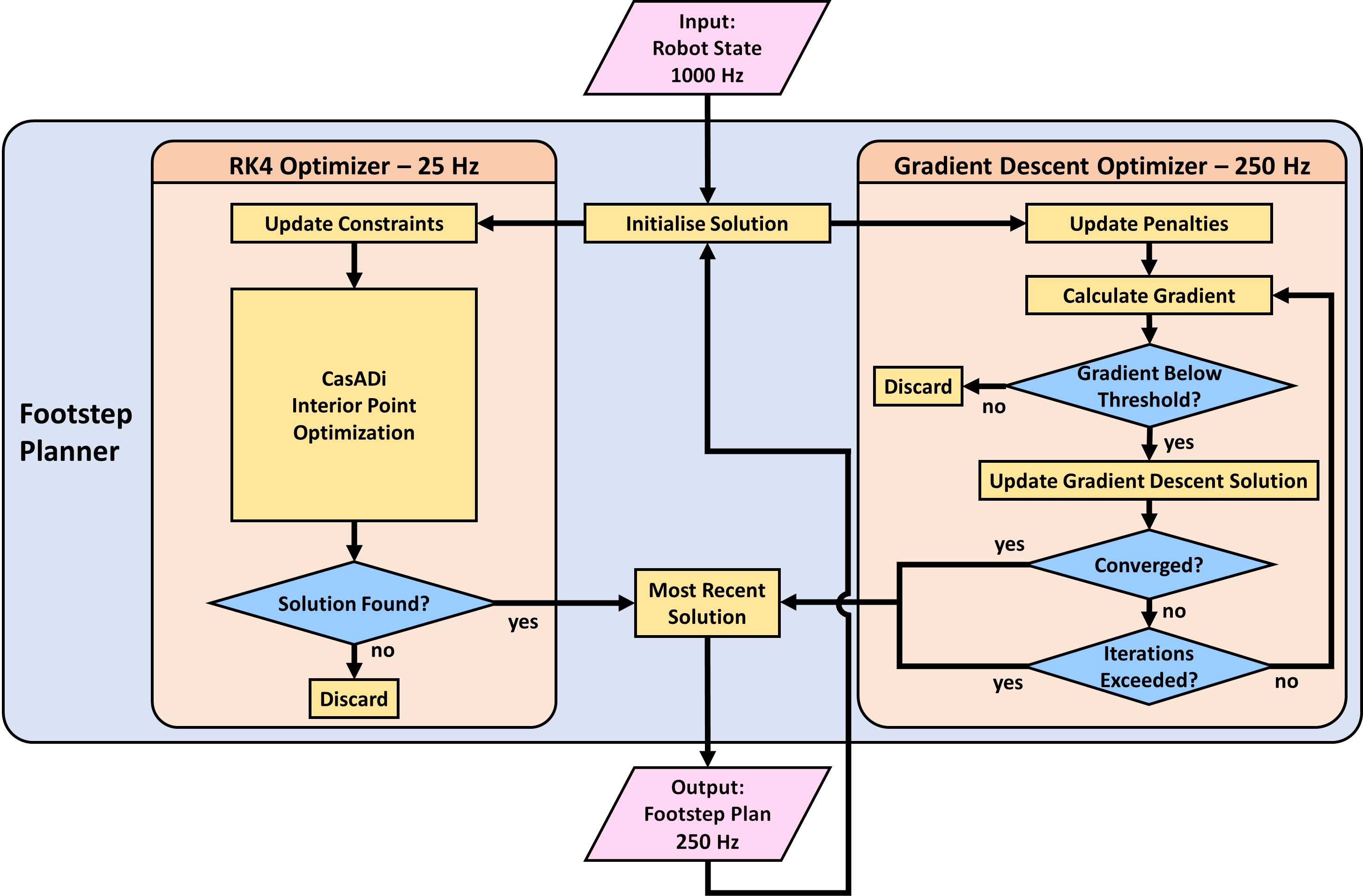}
    \caption{Flowchart describing the asynchronous optimization process.}
    \vspace{-0.5cm}
    \label{fig:Parallel}
\end{figure*}

The RK4 and gradient descent optimizers run in parallel threads communicating via ROS topics. Importantly, if the analytical gradient is extremely large (for example at the switching point between feet), the gradient descent solution is discarded. For periods of time where a gradient descent solution is infeasible, this means the RK4 solution is used, maintaining a feasible solution until gradient descent becomes feasible again. The interaction between the solvers is shown in Fig. \ref{fig:Parallel}. This method allows gradient descent to interpolate between RK4 optimizer solutions at a rate of $250\,$\si{Hz}: an order of magnitude increase of speed.

The reason an asynchronous combination of the optimizers is chosen over a synchronous variant is to maximise optimization speed; as soon as an optimizer produces a solution it can be used, instead of waiting until the fixed time. It should be noted that in experiments we simulate this at a real time factor below $1$ to maintain a high simulation accuracy. To make sure the optimizers run at realistic speeds relative to the simulation, we limit the solutions of the RK4 optimizer and the gradient descent optimizer to $25\,$\si{Hz} and $250\,$\si{Hz} respectively.

To perform the nonlinear RK4 optimization the software framework CasADi \cite{Andresson2012} was used to interface with the IPOPT \cite{Wachter2006} interior point optimization solver. To warm start the solver, the initial value of the solution, $\bm{u}_{k,\,j}^{(i)}$, is set as the previous iteration's final solution, $\bm{u}_{k,\,j-1}^{(f)}$, for each step in the prediction horizon, $k \in [1, ..., N]$. The current step duration, $\Delta t_{1,\,j}^{(i)}$ is adjusted for the time elapsed between iterations, $\tau_{j-1}$:
\begin{equation}
    \bm{u}_{k,\,j}^{(i)} = \bm{u}_{k,\,j-1}^{(f)}
\end{equation}
\begin{equation}
    \Delta t_{1,\,j}^{(i)} = \Delta t_{1,\, j-1}^{(f)} - \tau_{j-1}
\end{equation}
When a step has been taken between iterations, the initial value of the solution is provided as follows for the first $k \in [1, ..., N-1]$ predicted steps:
\begin{equation}
    \bm{u}_{k,\, j}^{(i)} = \bm{u}_{k+1,\, j-1}^{(f)}
\end{equation}
\begin{equation}
    \bm{\Delta t}_{k,\,j}^{(i)} = \bm{\Delta t}_{k+1,\, j-1}^{(f)}
\end{equation}
With the final predicted step initialized to repeat the motion of the penultimate step, mirrored in the $y$ direction.
\begin{equation}
    \bm{u}_{N,\, j}^{(i)} = \bm{u}_{N-1,\, j-1}^{(i)} + 
    \begin{bmatrix}
        1   & 0     \\
        0   & -1    \\
    \end{bmatrix}
    ( \bm{u}_{N,\, j-1}^{(i)} - \bm{u}_{N-1,\, j-1}^{(i)} )
\end{equation}
\begin{equation}
    \Delta t_{N,\,j}^{(i)} =  \Delta t_{N,\,j-1}^{(i)}
\end{equation}

Gradient descent optimization was directly implemented in C++, and runs for a maximum of $100$ steps. Because each footstep in the prediction horizon is dependent on the previous footstep, chain rule can be used to heavily speed up calculation and each individual derivative can be pre-calculated at the start of each gradient descent step. The initial value of the solution is initialized with the same method as the RK4 solver. The RK4 and gradient descent optimizers are running in parallel, publishing their respective solutions over ROS topics. The whole body controller runs at a higher rate of 1000 HZ using the QP solver qpOASES. The average solving time of one whole body control step is 750 microseconds.

\subsection{Trajectory Generation}

The swing foot trajectory is generated using fifth order polynomials: current and final positions, velocities and accelerations are specified, and a parametric quintic curve in the $x$, $y$, and $z$ directions generated to produce smooth trajectories. The $z$ polynomial is generated in two halves, with a midpoint foot height with zero velocity and acceleration specified to ensure no collision with the ground. The LIP model is used to generate the CoM trajectory during footsteps. Each trajectory point is calculated at the whole body controller rate of $1000\,$\si{Hz}.

\subsection{Whole Body Controller}
We use an inverse dynamics based whole body controller to track the desired CoM and swing foot trajectories. The task can be formulated as solving a QP without future prediction at each time step, similar to \cite{apgar2018fast}. The controller gives the optimal joint accelerations, contact forces and joint torques.
\begin{equation}
    \min_{\bm{\tau}, \bm{\ddot{q}}, \bm{f}} ||\dot{J}\bm{\dot{q}} + J\bm{\ddot{q}} -\bm{\ddot{x}}_{cmd}||_{W}
    \label{eq:WBC}
\end{equation}
 s.t. 
\begin{equation}
    M(q)\bm{\ddot{q}} + C(q. \dot{q}) + G(q) = S^{T}_{a}\bm{\tau} + J^{T}_{c}\bm{f}
    \label{eq:WBC_constrain1}
\end{equation}
\begin{equation}
    \bm{f}\in{P}
    \label{eq:WBC_constrain2}
\end{equation}
\begin{equation}
    S_{f}\bm{f} = 0
    \label{eq:WBC_constrain3}
\end{equation}
\begin{equation}
 \bm{\underline{\tau}} \leq \bm{\tau} \leq \bm{\Bar{\tau}} 
 \label{eq:WBC_constrain4}
\end{equation}
The commanded acceleration $\bm{\ddot{x}}_{cmd}$ is computed by PD control plus the feedforward term:
\begin{equation}
    \bm{\ddot{x}}_{cmd} = K_{p}(\bm{x}-\bm{x}_{d}) + K_{d}(\bm{\dot{x}} - \bm{\dot{x}}_{d}) + \bm{\ddot{x}}_{d}
    \label{eq:WBC_cmd}
\end{equation}
where $\bm{x}^{T} = [\bm{\theta}^{T}_{com} \hspace{0.35cm}
 \bm{x}^{T}_{com} \hspace{0.35cm}
 \bm{x}^{T}_{left} \hspace{0.35cm}
 \bm{x}^{T}_{right}] \in \mathbb{R}^{1 \times 12}$ comprises the actual orientation and position of the CoM, and actual positions of the left and right feet. The objective function in (\ref{eq:WBC}) simply minimizes the weighted error between actual and desired acceleration. Equation (\ref{eq:WBC_constrain1}) is the rigid body dynamics constraint. Equation (\ref{eq:WBC_constrain2}) constrains non-slippery contact by using  a pyramidal friction model, 
 \begin{equation}
     P = \{ (f_{x},f_{y},f_{z}) \in \mathbb{R}^{3}| f_{z} \geq	0; |f_{x}|, |f_{y}| \leq \frac{\mu}{\sqrt{2}}f_{z} \}
 \end{equation}
 When one foot is in the air, the force selection matrix $S_{f}$ in equation (\ref{eq:WBC_constrain3}) is used to constrain the forces on the foot to be 0. In equation (\ref{eq:WBC_constrain4}) torque limits are constrained. 
\section{EXPERIMENTAL SETUP}

\subsection{Walking Robot Platform: SLIDER}

To perform experiments, the walking robot SLIDER \cite{Wang2020} was used. SLIDER is a novel walking robot; each leg is actuated with a prismatic joint at the hip, which can also rotate in the roll and pitch axes. Because of this novel design, SLIDER has ultra-light legs, bringing it closer to the LIP walking model used during footstep planning. Experiments were carried out in simulation using Gazebo; similar to the real system (see Fig. \ref{fig:SLIDER}), the simulated SLIDER  has four force sensors on each foot to measure centre of pressure. Each ankle has two degrees of freedom: pitch and roll, giving SLIDER a total of ten degrees of freedom (five per leg). Importantly, SLIDER is not equipped with an upper body, meaning internal angular momentum cannot be used to stabilise or accelerate itself. SLIDER has a mass of only $15\,$\si{kg}, and during experiments, body height is maintained at $0.8\,$\si{m}.

To test the capabilities of ARTO for footstep planning, three experiments were performed to compare against standard MPC, which optimizes only footstep location at a rate of $500\,$\si{Hz} using quadratic programming, and against the RK4 optimizer only, which optimizes footstep location and duration at a rate of $25\,$\si{Hz}.

\subsection{Experiment 1: Push Recovery Analysis}

To qualitatively test the push recovery capability of the footstep planner, impulses were applied to the centre of mass of the robot by applying increasing forces for a fixed duration of $0.1\,$\si{s}. To enable direct comparison between footstep planning algorithms, the simulation was initialized to the same state each time, and the magnitude and time of force application was identical for each run.

\subsection{Experiment 2: Maximum Push Recovery}

Inspired by \cite{Khadiv2020}, the maximum push force was found for varying angles parallel to the $x-y$ (ground) plane. Pushes were applied for a fixed duration of $0.1\,$\si{s}.

\subsection{Experiment 3: Reference Velocity Changes}

To test the capability of the footstep planner for agile locomotion, the maximum velocity reference change from walking in place was found for varying angle of reference velocity. In a sense, this represents the maximum acceleration the planner can handle in each direction. It is important to note that this is to find the maximum reference change that the planner can handle while maintaining stability in the robot, rather than the maximum reference velocity that the robot can reach.

\begin{figure}[t]
    \centering
    \vspace{0.5cm}
    \includegraphics[width=0.45\textwidth]{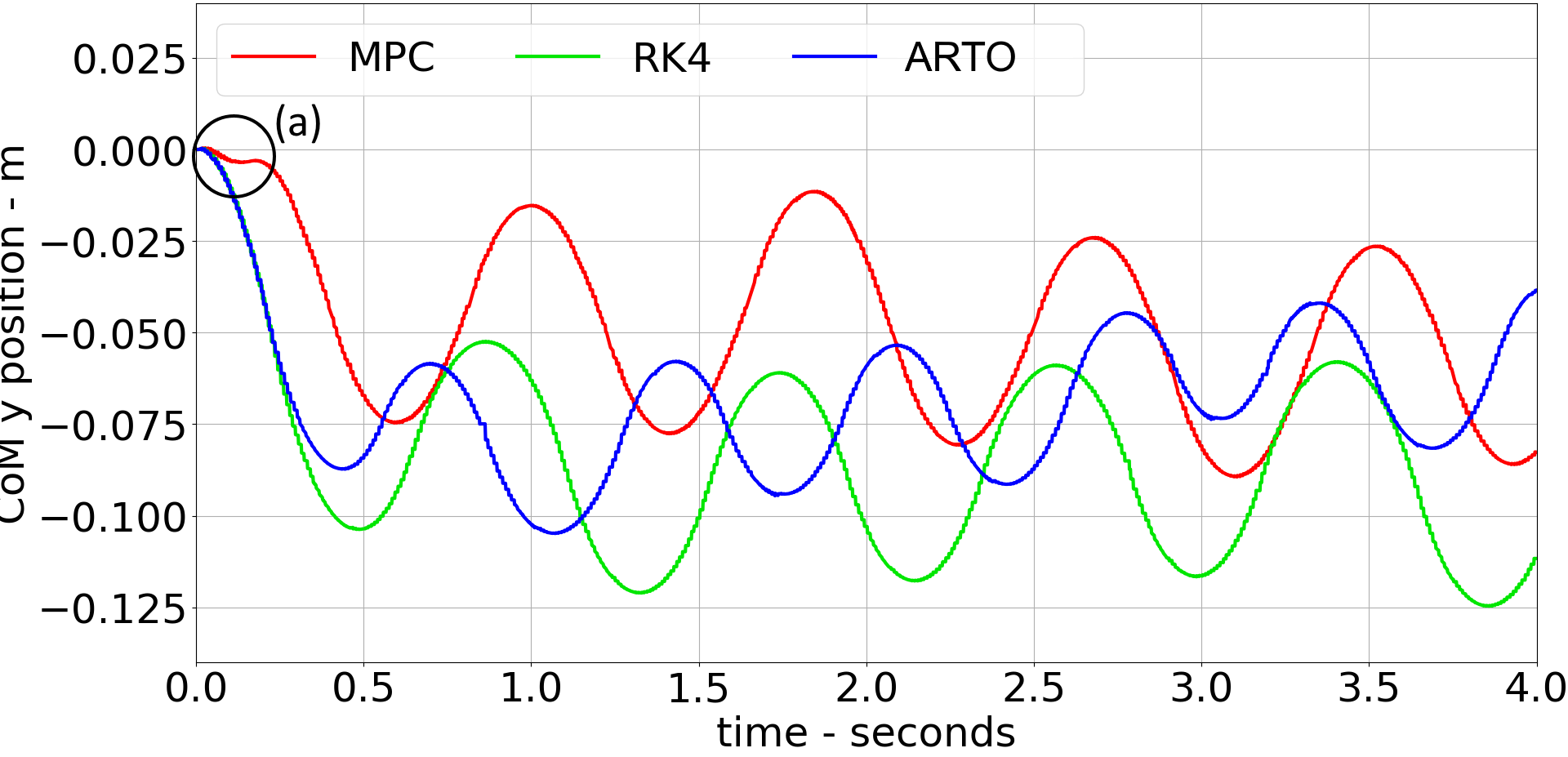}
    \includegraphics[width=0.45\textwidth]{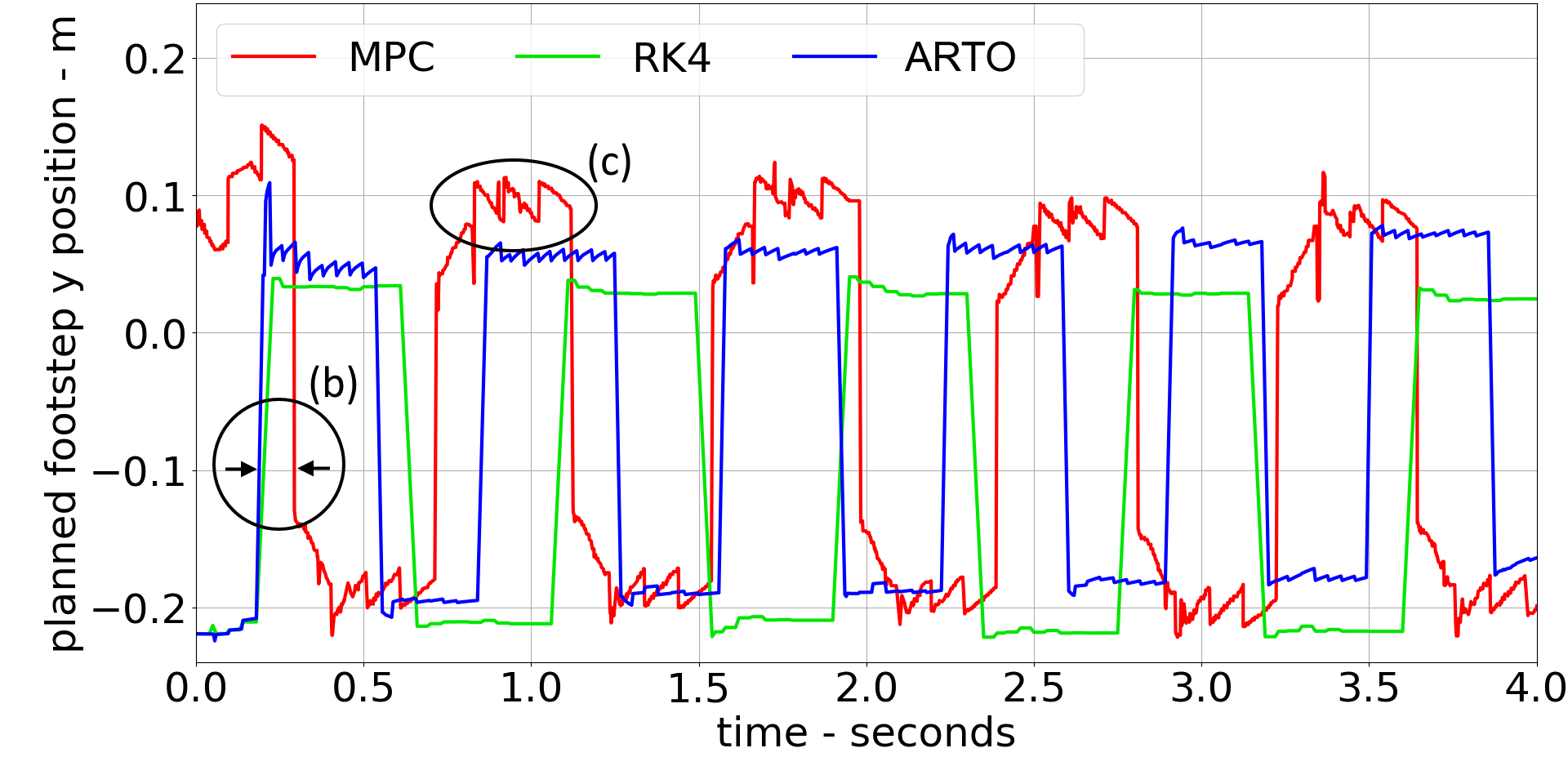}
    \caption{CoM $y$ position (top) and planned footstep $y$ position (bottom) with a reference velocity of $\bm{x}_{ref}=[0.1,\, 0]^T\,$\si{m.s^{-1}}, starting from stationary. Red: MPC, green: RK4, blue: ARTO. Detail $(a)$: assisted start for MPC, $(b)$: RK4 and ARTO produce shorter duration first steps, $(c)$: MPC produces greater variation in step location plans.}
    \vspace{-0.5cm}
    \label{fig:PlanTime}
\end{figure}

\section{RESULTS \& DISCUSSION}

The results of preliminary experiments show that the ARTO footstep planner produces valid footstep plans. The planned footstep $y$ position and remaining step duration for a location only MPC, RK4, and ARTO planners starting from stationary with a reference velocity of $\bm{x}_{ref}=[0.1,\, 0]^T\,$\si{m.s^{-1}} are shown in Fig. \ref{fig:PlanTime}. The standard MPC location optimizer was unable to start from stationary without an assisting force of $40\,$\si{N} for $0.1\,$\si{s} preventing the CoM from accelerating to the point of instability (seen in Fig. \ref{fig:PlanTime}, detail $(a)$). As seen, RK4 and ARTO settle on similar footstep timing and positions, and manage to reach a stable walking pattern after one initial step. Importantly, the first step transitioning from stationary to walking is considerably shorter using the RK4 and ARTO planners; seen in detail $(b)$. Standard MPC with assistance reaches a similar walking pattern when tuned with similar step durations, but the planned $y$ location of the foot is strongly discontinuous, as shown in detail $(c)$. The optimal footstep duration during stable walking found by the RK4 and ARTO planners is shorter than the fixed duration MPC planner; the step duration optimization has an effect.

\begin{figure}[t]
    \centering
    \vspace{0.5cm}
    \includegraphics[width=0.45\textwidth]{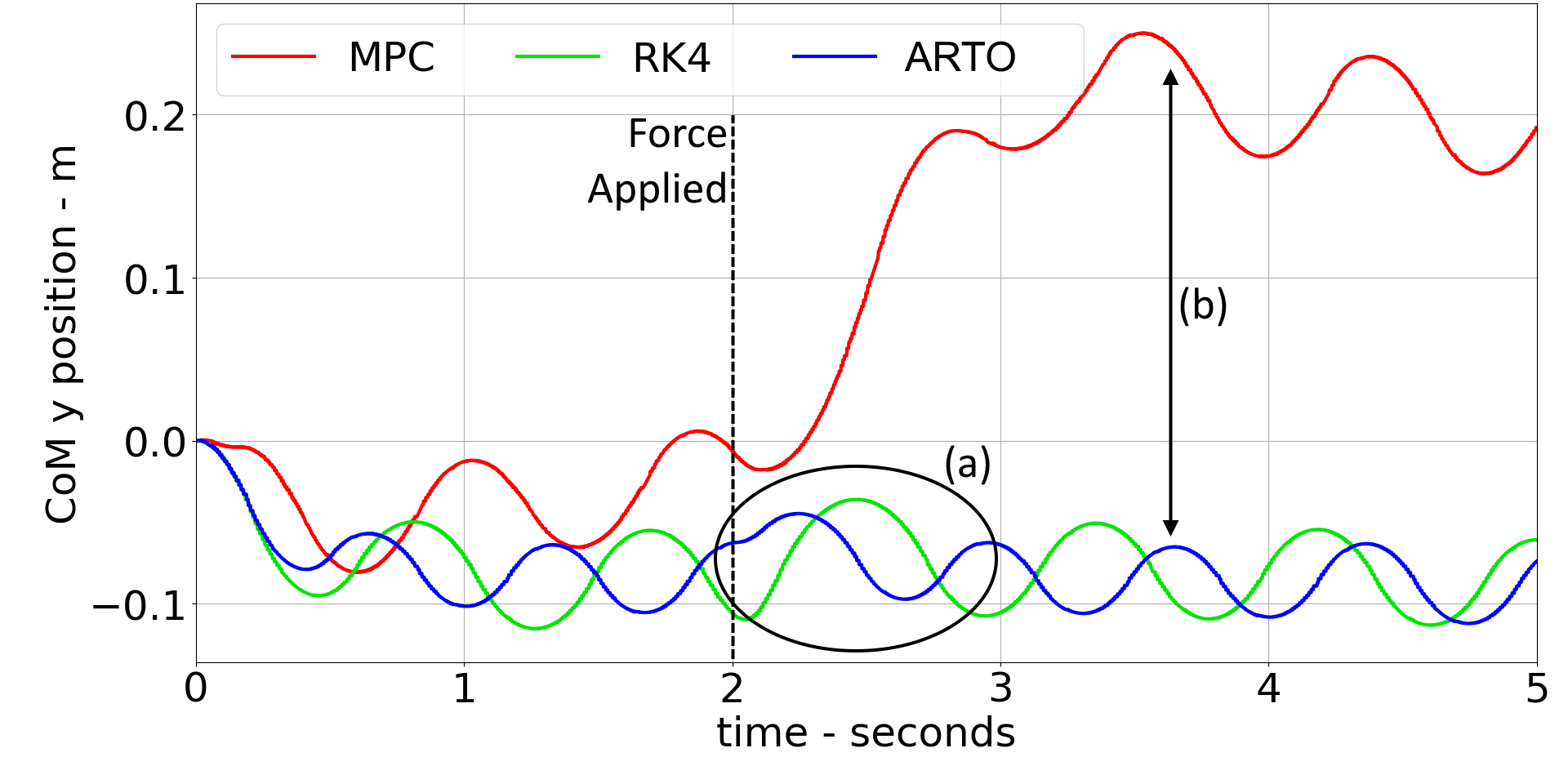}
    \includegraphics[width=0.45\textwidth]{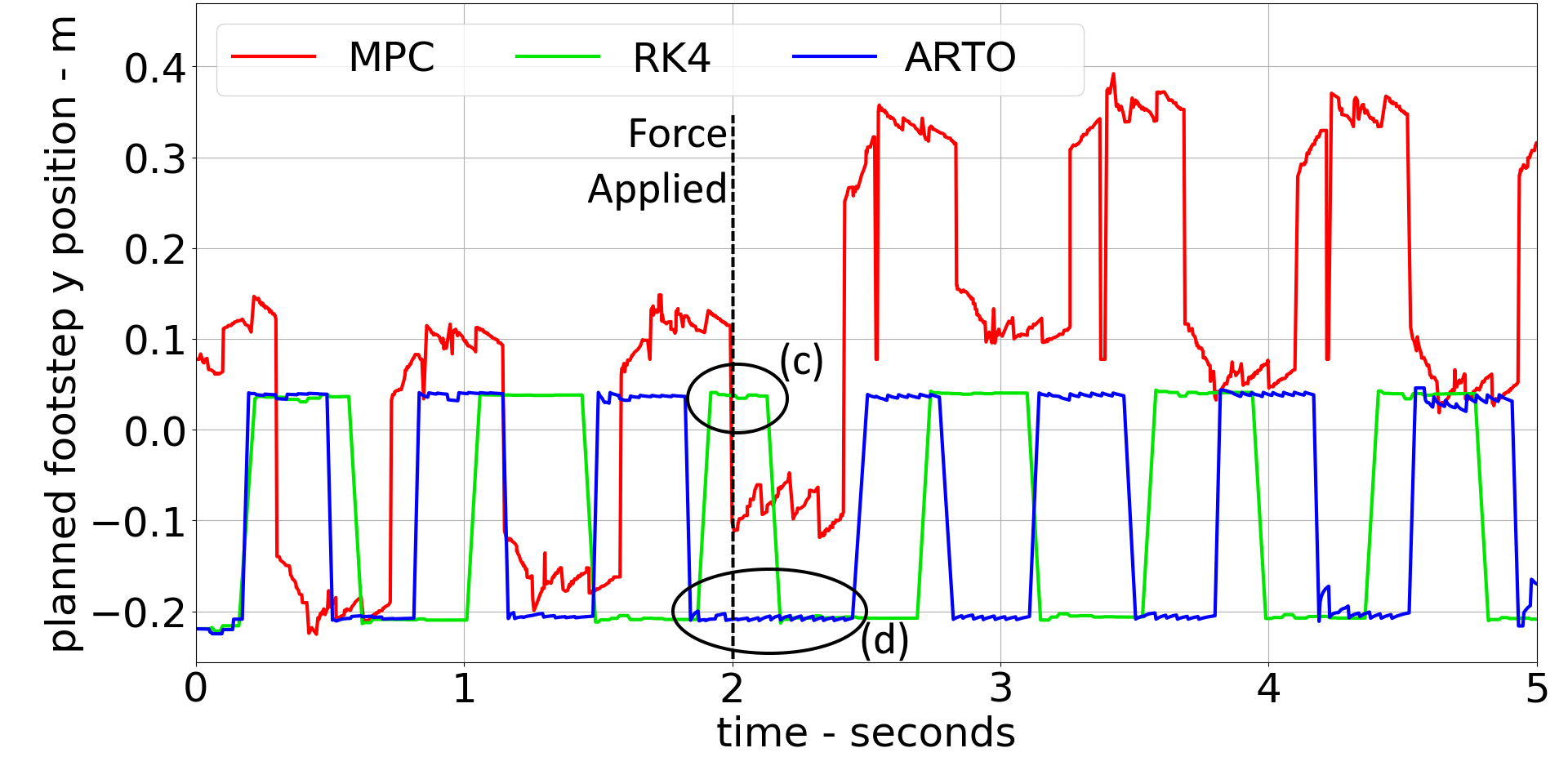}
    \caption{CoM $y$ position (top) and planned footstep $y$ position (bottom) when stepping in place, starting from stationary. Impulse is applied in the $y$ direction at $T=2.0\,$\si{s}.. Red: MPC, green: RK4, blue: ARTO. Detail $(a)$: RK4 and ARTO adapt to the push without stepping laterally, $(b)$: MPC steps laterally to compensate for the push, $(c)$: RK4 shortens the length of its current step to quickly step to the resisting foot, $(d)$: ARTO lengthens its current step to remain on the resisting foot}
    \vspace{-0.5cm}
    \label{fig:PushTime}
\end{figure}

With the robot stepping in place, a force of $40\,$\si{N} was applied to the CoM in the $y$ direction for $0.1\,$\si{s} at $2$ seconds into the simulation. RK4 and ARTO both resist the sideways push without adjusting step positions, as seen in Fig. \ref{fig:PushTime}, instead increasing the duration of the resisting step. Details $(a)$ and $(b)$ highlight how RK4 and ARTO manage to recover from the push within two steps and maintain the CoM trajectory with little deviation, while MPC takes multiple steps to recover and the CoM trajectory changes significantly in the direction of the applied force. Details $(c)$ and $(d)$ highlight the two possible strategies that optimizing footstep duration can use to resist the push; RK4 shortens its current step to shift weight onto the foot that can resist the push, while ARTO extends the duration of its current step to resist the push.

\begin{figure}[t]
    \centering
    \vspace{0.3cm}
    \includegraphics[width=0.43\textwidth]{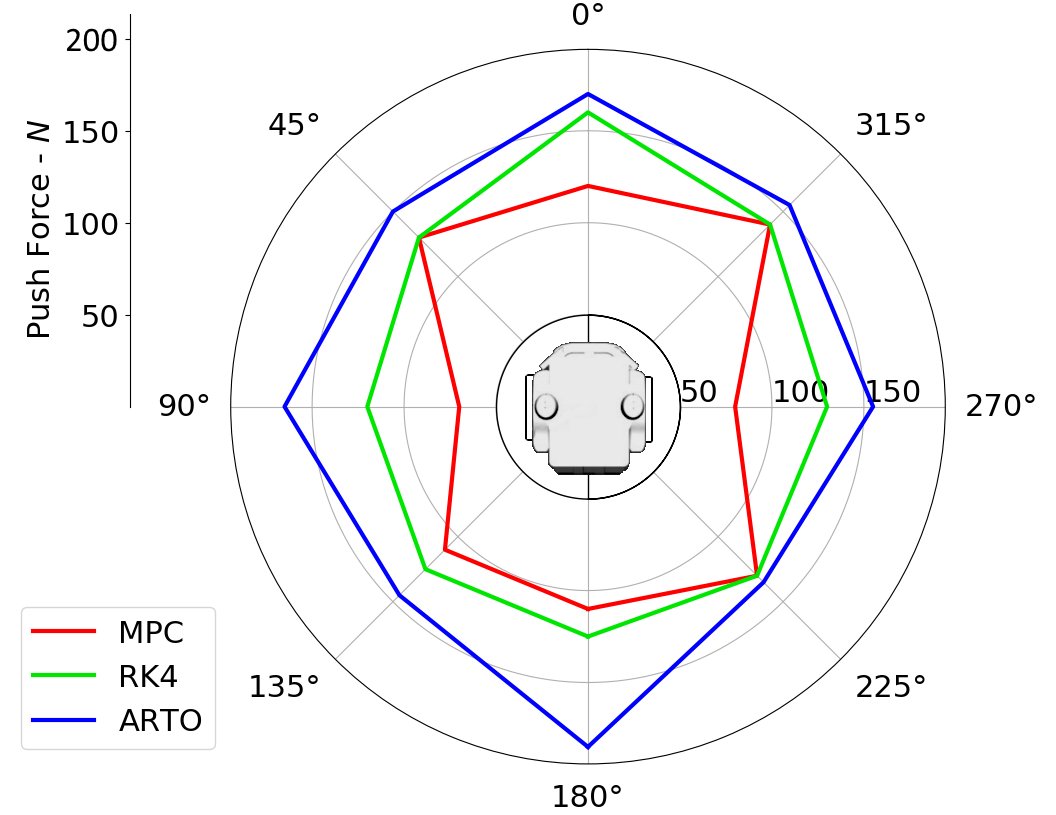}
    \caption{The maximum push force that can be recovered from in each direction relative to the front of the robot. Force applied for $0.1 \,$\si{s} to robot stepping in place. Red: MPC, green: RK4, blue: ARTO.}
    \vspace{-0.5cm}
    \label{fig:MaxPush}
\end{figure}

Figure \ref{fig:MaxPush} shows that optimizing footstep duration greatly improves the capabilities of RK4 and ARTO to resist sideways pushes compared to standard MPC. This is because constraints to prevent foot collision (\ref{eq:foot_constraints}) on footstep position limit location-only optimization. Further to this, it can be seen that ARTO outperforms RK4 and standard MPC when pushed from any direction. This is because ARTO can produce solutions an order of magnitude faster than RK4, allowing it to replan footsteps quick enough to deal with recovering from the edge of instability. Alongside faster optimization, ARTO also has redundancy in optimization solutions; if the RK4 optimizer fails to produce a feasible solution, the gradient descent optimizer can operate independently, and vice versa.

Experiments testing the ability of each planner to handle reference velocity changes support each point from the push recovery experiments. The maximum sideways reference velocity change for each planner is lower than forwards or backward velocity changes, primarily because of footstep collision avoiding constraints. Even at low optimization frequency, RK4 generally outperforms standard MPC optimization, indicating that optimizing footstep duration elevates the robots capacity to handle larger reference changes. ARTO improves upon this further in all directions; redundancy in optimization solutions increases the probability that feasible footstep plans are found for a given reference change.

\begin{figure}[t]
    \centering
    \vspace{0.5cm}
    \includegraphics[width=0.43\textwidth]{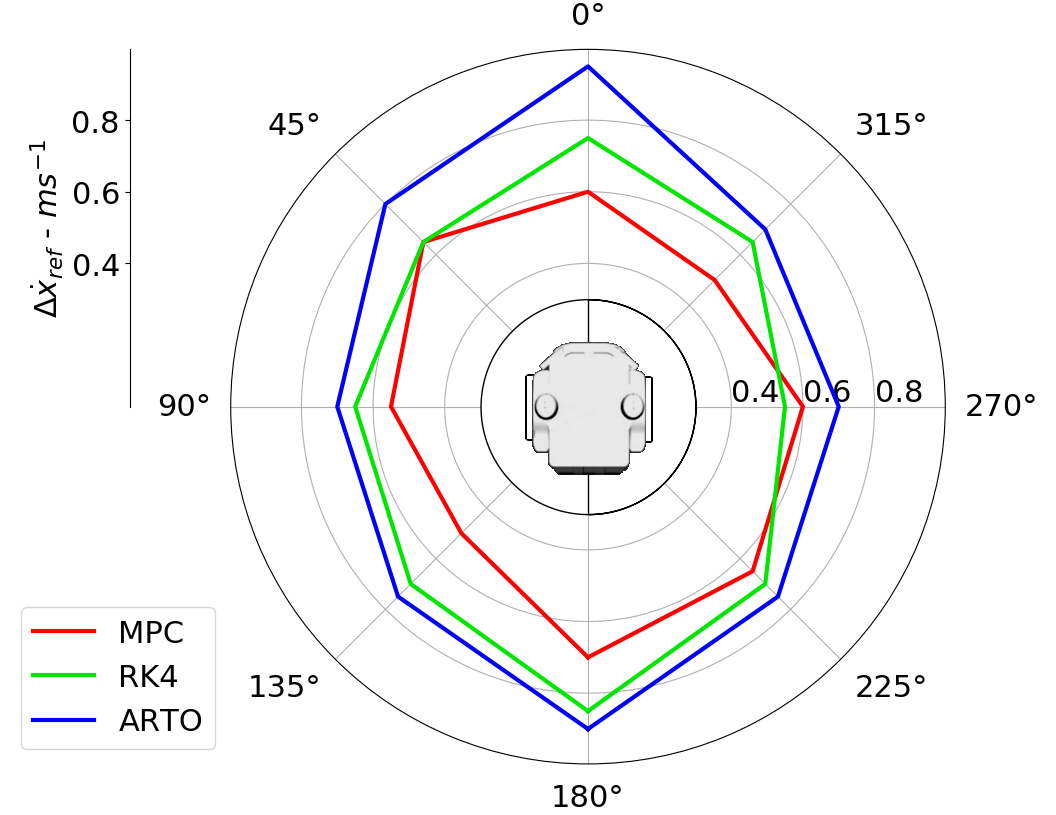}
    \caption{The maximum reference velocity change from stepping in place in each direction relative to the front of the robot. Red: MPC, green: RK4, blue: ARTO.}
    \vspace{-0.5cm}
    \label{fig:MaxV}
\end{figure}

\section{CONCLUSIONS \& FUTURE WORK}

The proposed footstep planner offers a number of benefits over traditional location-based planners, both qualitative and quantitative. Qualitatively, both the RK4 and ARTO planners were able to begin walking from a stationary start, while traditional MPC that optimizes footstep location requires either starting assistance or a double support phase starting motion. It was shown that footstep planners that performed time adaptation were able to resist a push without compromising step location; something that is extremely useful for constrained environments such as restricted work spaces. Quantitatively, it was shown that using gradient descent to interpolate between RK4 plans allows the footstep planner to recover from larger pushes, particularly in the lateral direction. Further to this, ARTO was able to adjust to larger velocity reference changes than RK4 or MPC planners, utilizing redundancy in optimization solutions when having two optimizers running in parallel to find feasible footstep plans for each reference change. MPC was shown to react less effectively than both RK4 and ARTO to reference velocity changes, indicating that step timing is vital to accelerate and decelerate effectively.

The experiments conducted in this paper are useful in two scenarios; reference velocity, and external forces. By controlling the reference velocity of the walking robot, it would be possible to direct the robot while the lower level footstep planner ensured the robot maintained stability. Applying external forces on the robot, for example if an extra upper mass was displaced in one direction, the walking robot would step to counteract this, thereby walking in the direction of the force. Beyond bipedal walking robots, it is hoped that ARTO can be applied to other scenarios where gradient descent can lend itself well to the optimization problem, like quadruped robots.

\addtolength{\textheight}{-0.5cm}   


\section*{APPENDIX}
\subsection{Analytical Gradients}
\subsubsection{Footstep position gradients}
\begin{multline*}
    \frac{\partial x_j}{\partial u_{x, k}} =
    \begin{cases}
        0 & j \leq k\\
        -\frac{1}{2}(e^{\omega \Delta t_{j-1}} + e^{-\omega \Delta t_{j-1}}) + 1 & j = k + 1\\\\
        \frac{1}{2}\frac{\partial x_{j-1}}{\partial u_{x, k}}(e^{\omega \Delta t_{j-1}} + e^{-\omega \Delta t_{j-1}}) +& j \geq k+2\\
        \frac{1}{2\omega}\frac{\partial \dot{x}_{j-1}}{\partial u_{x, k}}(e^{\omega \Delta t_{j-1}} - e^{-\omega \Delta t_{j-1}})
    \end{cases}
\end{multline*}

\begin{multline*}
    \frac{\partial \dot{x}_j}{\partial u_{x, k}} =
    \begin{cases}
        0 & j \leq k\\
        -\frac{1}{2}\omega(e^{\omega \Delta t_{j-1}} - e^{-\omega \Delta t_{j-1}}) & j = k + 1\\\\
        \frac{\omega}{2}\frac{\partial x_{j-1}}{\partial u_{x, k}}(e^{\omega \Delta t_{j-1}} - e^{-\omega \Delta t_{j-1}}) + & j \geq k+2
        \\\frac{1}{2}\frac{\partial \dot{x}_{j-1}}{\partial u_{x, k}}(e^{\omega \Delta t_{j-1}} + e^{-\omega \Delta t_{j-1}})
    \end{cases}
\end{multline*}
With identical formulation in the $y$ direction.
\subsubsection{Footstep duration gradients}
\begin{multline*}
    \frac{\partial x_j}{\partial \Delta t_k} =
    \begin{cases}
        0 & j < k
        \\
        \dot{x}_j & j=k
        \\
        \frac{1}{2}\frac{\partial x_{j-1}}{\partial \Delta t_{k}}(e^{\omega \Delta t_j} + e^{-\omega \Delta t_j}) + & j \geq k+1
        \\
        \frac{1}{2 \omega}\frac{\partial \dot{x}_{j-1}}{\partial \Delta t_{k}}(e^{\omega \Delta t_j} - e^{-\omega \Delta t_j})
    \end{cases}
\end{multline*}\vspace{-0.5cm}
\begin{multline*}
    \frac{\partial \dot{x}_j}{\partial \Delta t_k} =
    \begin{cases}
        0 & j < k
        \\
        \ddot{x}_j & j=k
        \\
        \frac{\omega}{2}\frac{\partial x_{j-1}}{\partial \Delta t_{k}}(e^{\omega \Delta t_j} - e^{-\omega \Delta t_j}) + & j \geq k+1
        \\
        \frac{1}{2}\frac{\partial \dot{x}_{j-1}}{\partial \Delta t_{k}}(e^{\omega \Delta t_j} + e^{-\omega \Delta t_j})
    \end{cases}
\end{multline*}
Again, with identical formulation in the $y$ direction.
\\
\section*{ACKNOWLEDGMENT}

This work was supported by UK Research and Innovation [grant number EP/S023283/1]. Digby Chappell is funded by the UKRI Centre for Doctoral Training in AI for Healthcare. Ke Wang is funded by CSC Imperial Scholarship.
\\
\bibliographystyle{ieeetr}
\bibliography{bibliography.bib}

\end{document}